# Analysis of Real-Time Hostile Activitiy Detection from Spatiotemporal Features Using Time Distributed Deep CNNs, RNNs and Attention-Based Mechanisms

Labib Ahmed Siddique[1], Rabita Junhai[2], Tanzim Reza[3], Salman Sayeed Khan[4] and Tanvir Rahman[5]
[1,2,3,4]Department of Computer Science and Engineering, BRAC University, 66 Mohakhali, Dhaka 1212, Bangladesh
[5]Department of Computer and Information Sciences, College of Engineering, University of Delaware
Email: [1]labib.ahmed.siddique@g.bracu.ac.bd, [2]rabita.junhai@g.bracu.ac.bd,
[3]tanzim.reza@bracu.ac.bd, [4]salman.sayeed@bracu.ac.bd, [5]rtanvir@udel.edu,

*Abstract*—Real-time video surveillance, through CCTV camera systems has become essential for ensuring public safety which is a priority today. Although CCTV cameras help a lot in increasing security, these systems require constant human interaction and monitoring. To eradicate this issue, intelligent surveillance systems can be built using deep learning video classification techniques that can help us automate surveillance systems to detect violence as it happens. In this research, we explore deep learning video classification techniques to detect violence as they are happening. Traditional image classification techniques fall short when it comes to classifying videos as they attempt to classify each frame separately for which the predictions start to flicker. Therefore, many researchers are coming up with video classification techniques that consider spatiotemporal features while classifying. However, deploying these deep learning models with methods such as skeleton points obtained through pose estimation and optical flow obtained through depth sensors, are not always practical in an IoT environment. Although these techniques ensure a higher accuracy score, they are computationally heavier. Keeping these constraints in mind, we experimented with various video classification and action recognition techniques such as ConvLSTM, LRCN (with both custom CNN layers and VGG-16 as feature extractor) CNNTransformer and C3D. We achieved a test accuracy of 80% on ConvLSTM, 83.33% on CNN-BiLSTM, 70% on VGG16-BiLstm ,76.76% on CNN-Transformer and 80% on C3D.

## I. INTRODUCTION

Globalization resulted in a more advanced world with cutting-edge technologies and innovations. Along with advancements, it also led to an increase in criminal activity around the world. This made crime detection and its prevention a dire need in today's time. Video surveillance systems proved to be an efficient method for monitoring crimes as they can record human activities in real-time. Installing a camera in a location acts as an eye that monitors the area and helps to provide security to those who are within its range. The goal of surveillance systems deployed in schools, hospitals, parks, prisons, banks, markets, streets, etc. is to detect hostile activities and alert the concerning authorities about the situation. These systems give crucial data gathered straight from the action scene. However, the volume of video recordings might quickly overwhelm human operators. Distinguishing between violent and non-violent scenes at times becomes difficult and time-consuming, resulting in little to no manual response at all. Moreover, activities such as jogging, dancing, or face-to-face conversations, appear to be extremely comparable to aggressive conduct. As a result, significant research effort was dedicated to inventing systems that proactively interprets data in an attempt to detect anomalous behavior, alert automatically, and securely delete unnecessary in- formation. Although these researches do not include activities such as fist bumps, high fives, hugs, and so on which results in false positive predictions. A variety of Deep Learning algorithms are developed to study human actions in real life. HAR is a popular time series classification problem in which data from multiple time step is used to accurately classify the actions being performed. Moreover, for video classification tasks, developing an image classifier will not be feasible as it tries to predict every single frame, resulting in prediction flickering. On top of that, image classifiers do not take environmental context into consideration. Therefore, keeping these constraints in mind, we proposed deep learning models that can be easily integrated with IOT such as LRCN (both custom CNN and pre-trained), C3D, ConvLSTM, and CNN-Transformers to help automate surveillance systems. These models when deployed, are trained to have high accuracy and help classify non-violent behaviors from violent ones. The key contribution to the research are these models are lightweight and easily deployable, trained on real life gestures that can result on false positive alarms and have lower inference time.

The primary objectives of this research are as follows:

- To create deep learning models to detect violent or non-

- violent human behavior in an automated way.
- To process raw CCTV footage to extract features so that it can help us classify normal human activity from abnormal human activity.
- To make a real-time monitoring system with the help of IoT that will be cost-efficient, safe, and more effective than the existing surveillance systems.

## II. RELATED WORKS:

In [1], a deep learning classifier was proposed to detect abnormal crowd activities. To extract the features of videos collected from movies, researchers used Fully Convolutional Neural Networks (FCN). A pre-trained FCN was fed the optical flow of two successive video frames as well as the individual frames, derived from AlexNet to obtain more valuable appearance and motion information. Their method provided both spatial and temporal continuity. They replaced the two-stream CNN architecture with a two-stream FCN one . They designed a simple but effective method for encoding highdimensional feature maps and then used binary codes to find patterns. To determine the degree of abnormality in the video, the abnormal coefficient was calculated using an iterative quantization (ITQ) method based on the feature map from the FCN. In paper [2], the researchers presented a low latency human detector for unmanned aerial vehicles (UAVs) with optical flow and CNNs. The proposed method included quick ROI generation and extraction and a two-stream CNN classifier to detect running people by distinguishing appearance and motion features from walking people or other interferences. They came up with ROIs for human categorization in real time. The optical flow was calculated with every two successive frames to locate the candidate targets quickly. A series of preprocessing techniques were used to extract the ROIs, Including morphological expansion, spatial average filtering, and outer contour extraction. A small-kernel CNN was presented to accurately recognize running humans in varied backgrounds. The small kernel reduced inference time. Field experiments and benchmark testing revealed that their system could recognize moving people at 15 FPS with an accuracy of 81.1% in complex environments for UAV scenes.

Two Stacked Denoising Autoencoder (SDAE) network was used to learn appearance and motion features from videos to reduce computational complexity in [3]. The first SDAE was used to capture static appearance clues. The second SDAE extracted motion features. Using two SDAEs, they extracted depth appearance features and depth motion features of the trajectories. They used the bag of word method to build two vocabularies. At first, they extracted deep motion features, which were then clustered. To get the most compact vocabulary, the Agglomerative Information Bottleneck approach was used. To reduce the vocabulary and minimize mutual information loss, their approach repeatedly combined two visual terms. Adaptive feature fusion methods improved the distinctiveness of these features. The researchers used Sparse representations to detect abnormal behaviour and improve detection accuracy. In [4], they proposed a model which was able to detect abnormal activity without the presence of an alarming or harmful object itself. The model generated macroblock motion vectors from video compression methods with three purposes which were: requirement of reliability with low false-positive rate, the ability to distinguish between normal and abnormal activity, and identification of abnormal behavior with less computing power. They strictly restricted the use of segmentation and tracking to avoid any kind of semantic interpretation. So, they first converted the video to motion vectors by compressing the data. They used motion vectors to generate motion elements, and the entering frame was assessed to the predictive model, where low probability indicated abnormal activity. They used probability distribution of each frame to prepare a histogram and detected the frame with less occurrence of features below a threshold value as abnormal.

The paper [5] suggested a model that was used to detect any suspicious presence with the help of a message alert. Here, they included the background removal technique to improve the detection of the moving object. STIPs are retrieved from depth videos using a compression algorithm that successfully eliminates chaotic readings. The system looked at mainly three limitations like the height, time, and body movement therefore, when all the aspects will be satisfied the person will be identified as a doubtful person. Firstly, the background reduction was done on the raw video data. After which the extracted frame from the video was converted in-frame sequence. This helped them to recognize the actions and when suspicious activities were detected the alarm will go off sending alert messages. In paper[13], Single frame prediction methods were utilized to automate surveillance systems to identify unusual behavior. The research integrated transfer learning on models pre trained on imagenet to classify anomalous behaviors ranging from pickpocketing, burglary, breaking in, and so on. The paper [14] designed spatial-temporal CNN frameworks to generate high-level features from both spatial and temporal dimensions. The model used visual data from a single, static image as well as dynamic volume data from consecutive frames to identify and classify anomalous events in overcrowded surveillance videos. The model was only used on spatial-temporal volumes of interest (SVOI) to optimize operational cost. In another paper, [15] the Deep Belief Networks (DBNs) architecture was used to learn robust features for pattern recognition where an unstructured DBN was developed to retrieve generalized underlying characteristics and a one-class SVM was taught to use the DBN's features that learned to retrieve properties that were taught to compress high-dimensional inputs into a low-dimensional component set using a non-linear dimensionality reduction approach.

## III. WORK PLAN

Our research aims to create deep learning models that can classify CCTV video footage into violent or nonviolent. We aim to train the models to consider classifying the entire video on a specific time frame as opposed to every single frame separately. Single frame classifications result in flickering and cause high false alarms, so we explore techniques that will classify images in a sequence.

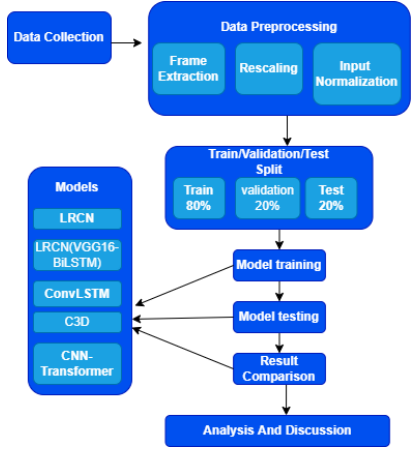

Fig. 1. The flow chart of the proposed hostile behavior detection model

## IV. METHODOLOGY

### A. Dataset Description

We used a research dataset created by Bianculi [6] et al. as our input data. In this dataset, there are 350 clips that are in MP4 format (H.24 codec) where 120 clips contain nonviolent videos and 230 clips contain violent videos. The frame rate per video is 30 fps, and the resolution is fixed to 1920x1080 pixels for all the clips.

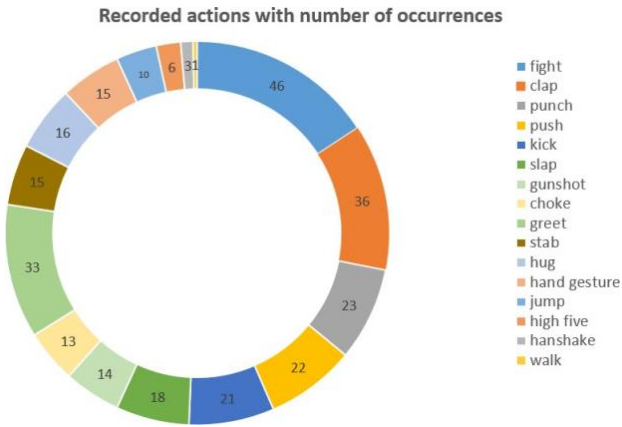

Fig. 2. This is the list of the recorded actions, with the number of occurrences in the dataset

The videos which are labeled as violent include generic hostile behaviors in public places whereas the nonviolent videos include normal gestures as well as the ones that are similar to the hostile ones. The average length of these videos is 5.63 seconds, where the largest is 14 seconds, and the shortest is 2 seconds. The characters in the videos are actors who acted out as per the requirements. In order to differentiate between the two behaviors, individuals were asked to indulge in actions that include kicking, slapping, punching, beating, firing guns, stabbing, hugging, clapping, being euphoric, exchanging high fives and waving at each other. These actions not only helped to detect abnormal behaviors but also prevented the results from being false positives which in turn assured more accurate results.

### B. Algorithm Description

*1) Long-term Recurent Convolutional Network:* The idea of LRCN [7] is to extract spatial features from each frame using convolution neural networks. The outputs of these convolutional networks are passes into a Bi-LSTM network, which fuses temporal features onto extracted spatial features to classify them. For our research, we built a custom CNN feature extractor model as well as experimented with a pre-trained model like VGG-16. Both the models take 90x90 pixels as input. For the custom CNN model, the convolutional layer is modeled to take the frames as inputs, perform operations using convolutional filters which are a matrix (in our case of 3x3 size) with a random set of values that convolve over the image and compute the dot operation and then pushes the output onto the next layer. The following equations (1,2,3,4) summarize an input frame and generate an output matrix by conducting convolution across $k$ channels:

$$A_o^{(m)} = g_m\left(\sum_k W_{ok}^{(m)} * A_k^{(m-1)} + b_o^{(m)}\right) \quad (1)$$

$$W-ok * A_k[s,t] = \sum_{p,q} a * b \quad (2)$$

$$a = A_k[s+p, t+q] \quad (3)$$

$$b = W_{ok}[P-1-p, Q-1-q] \quad (4)$$

Following each convolutional layer, max pooling, the number of parameters is shrunk in the network that cuts the convolutional load which is depicted in the preceding equations (5,6). The output is finally flattened. We used the following two activation functions in our model, Rectified Linear Unit (ReLU) [7] depicted in equation (7) and SoftMax [11] which converts a system's output into a probability distribution across expected classes. For the VGG-LSTM model, we used a VGG-16 network trained on ImageNet to extract features by importing the model and excluding the top. Then it was wrapped in a time distributed layer which was then passed to a bidirectional lstm of 256 filter followed by a dense layer of 256 filters with ReLu activation and the output layer with 2 neurons. The output vector from both the models is then passed to the time distributed layer which makes the models do the convolutional operations across a defined time-step so that the Bidirectional LSTM can learn the changes of the spatial features of the video frames and temporal weights across the defined time-step as well as learn the temporal changes better as all the hidden states contain information of both past and future. Finally, the output of the Bi-LSTM is passed to a dense layer with two heads at the end. The SoftMax activation function is utilized to replicate the 1-0 impulse carried away as an activation function. The layer has a bias parameter, b which is shown in the equations (8,9).

$$y = A_i \cdot x + b \quad (5)$$

$$y_i = \sum_{j=1} (A_{ij} x_j) + b_i \quad (6)$$

$$y = max(0, x) \quad (7)$$

$$y = A_i \cdot x + b \quad (8)$$

$$y_i = \sum_{j=1} (A_{ij} x_j) + b_i \quad (9)$$

*2) Convolutional Long Short Term Memory:* Since LSTM alone cannot deal with spatial data, we have also used ConvLSTM as it can be used for both the time and spectral domain. This is achieved by using ConvLSTM with 3D tensor inputs, cell outputs, hidden states and gates whose final dimensions are spatial. Although ConvLSTM has the same structure as LSTM, their main difference lies in the input-to-state and state-to-state transitions. In the following equations(10,11,12,13,14,15), the '$\sigma$', '' and '$\circ$' denotes the activation function, convolution operator and Hadamard product:

$$i_t = \sigma(w_{xi} x_t + w_{hi} h_{t-1} + w_{ci} \circ c_{t-1} + b_i) \quad (10)$$

$$f_t = \sigma(w_{xf} x_t + w_{hf} h_{t-1} + w_{cf} \circ c_{t-1} + b_f) \quad (11)$$

$$\tilde{c}_t = tanh(w_{xc} x_t + w_{hc} h_{t-1} + b_c) \quad (12)$$

$$c_t = f_t \circ c_{t-1} + i_t \circ \tilde{c}_t \quad (13)$$

$$o_t = \sigma(w_{xo} x_t + w_{ho} h_{t-1} + w_{co} \sigma c_t + b_o) \quad (14)$$

$$h_t = o_t \sigma tanh(c_t) \quad (15)$$

From the expressions above, we can say $C_{t-1}$ is the current position where $X_t$ is its input and $H_{t-1}$ is the state and result of the final neuron. The convolution filter is 2-dimensional with a $k \times k$ kernel where the dimension of the convolutional kernel is denoted by $k$. The ConvLSTM takes the frames of the video as the input and the multidimensional convolution operates on each frame to extract the features. Unlike the CNN model, ConvLSTM can transfer and process data in both, inter-layer as well as the intro-layer making it more efficient to extract features compared to CNN.

*3) 3D Convolutional Neural Networks:* The 3D Convolutional Neural Networks (C3D) [8] extract temporal and spatial features from video clips unlike the 2D–CNNs. This is because 2D convolution on a video segment squeezes the temporal features after convolving which results in an overall feature map with no dynamic depiction. In order to produce the 3D cube to obtain the 3D convolution, a 3D filter kernel is combined by stacking a number of frames together. The 3D CNN is designed in such a way that multiple feature maps can be generated in the later layers by placing them in the same location in the previous layers. The input dimensions for the videos are frames x height x width x channel in the manner: 25 x 90 x 90 x 3. The first 3D convolutional layer has 64 filters followed by a ReLu activation function. This is followed by a max pooling which estimates the highest value within every feature map patch and pools the most prominent feature in each patch. This is followed by another similar 3D convolutional layer as the first one with 32 filters with ReLu activation followed by another similar max pooling layer. The ConvNets extract the graphical properties of an image and organize them in a low-level representation such as a vector. The 3D CNN finds the vector for a stack of images to label the input video correctly. The flatten layers turn the input into a one dimensional vector output and pass it to the dense layers while adding weights to each data and classifying them followed by a dropout of 0.5.

*4) Convolutional Neural Network Transformer:* The Conv2D extracts spatial features, wraps them around a time distributed layer passes the output to the transformer. Transformers [12] use the attention-mechanism and are a sequence to sequence model, with an encoder and a decoder. Unlike other sequence models, they do not not use any Recurrent Neural Networks.

The structure of encoder and decoder allows the layering of modules several times on each other. The modules primarily consist of Feed Forward and Multi-Head Attention layers. In the embedded vector representation of each video, it is necessary to add the relative position. The equations (16, 17) define the process explained above where $V$, $Q$ and $K$ refers to the values, query and keys of all the videos in the sequence. The attention module consists of different video sequences for $V$ and $Q$ whereas the multi-head attention modele contains the same video sequence for both. The attention module does so by multiplying and adding the values in V with weights, called attention-weights as shown in equation (18):

$$Attention(Q, K, V) = softmax \frac{QK^T}{\sqrt{d_k}} V \quad (16)$$

$$a = softmax \frac{QK^T}{\sqrt{d_k}} V \quad (17)$$

The weights represent the influence of the video sequence of Q on that of K. The purpose of implying the SoftMax function is to create dissemination of 0 and 1. Repeating the attention mechanism several times enables the model to adapt and learn the various orientations of $V$, $Q$, and $K$. While training, the model learns the weight matrices, $W$, and multiplying them with $V$, $Q$ and $K$ results in the linear orientation of the video sequences. Each position of the attention module contains different matrices of $V$, $Q$ and $K$. This is because at a time, we can focus only on the entire input sequence of the encoder or a portion of input sequence of the decoder. The input sequence from the encoder and decoder is connected together up to a position by the multi-head attention module.

The next layer is the feed-forward which is a pointwise layer. This means that the network contains identical parameters at each point, and each video from the provided sequence is a distinct, identical linear transformation.

## V. IMPLEMENTATION AND RESULTS

### A. Implementation

To train our dataset in the desired models, we had to first preprocess the data. This step included extracting frames from each video, rescaling the frame, normalizing the input data and lastly, applying one-hot encoding. We extracted 25 frames per video to get a good training accuracy and rescaled them to ensure the images are all of the same size. In the preprocessing step, we labelled the nonviolent data as 0 and violent as 1. After this step, the dataset was divided into three sets which are train set, test set and validation set. Finally, we embedded the dataset into our models which are C3D, ConvLSTM, CNN-Transformer and LRCN (CNN- BiLSTM, VGG-BiLSTM).

### B. Result Analysis

*1) Accuracy and Quality Scores:* Amongst all the models, the CNN-Transformer model performs the most optimally with 0.74 being the F1 score for non violent and 0.79 being for the violent case.

| Model | Train Accuracy | Validation Accuracy | Test Accuracy | Train Loss | Validation Loss | Test Loss |
|---|---|---|---|---|---|---|
| ConvLSTM | 100.00 | 81.84 | 80.00 | 0.002 | 0.36 | 1.28 |
| LRCN | 99.57 | 81.48 | 83.33 | 0.024 | 0.73 | 0.82 |
| VGG Bi-LSTM | 85.90 | 77.78 | 70.00 | 0.33 | 0.46 | 0.68 |
| C3D | 93.16 | 85.19 | 80.00 | 0.12 | 0.44 | 1.15 |
| CNN-Transfromer | 97.86 | 88.89 | 76.67 | 0.078 | 0.32 | 0.69 |

TABLE I
ACCURACY SCORES

| Model | Class | Precision | Recall | F1-score |
|---|---|---|---|---|
| ConvLSTM | Non-Violent | 0.55 | 0.75 | 0.63 |
| | Violent | 0.89 | 0.77 | 0.83 |
| CNN Bi-LSTM | Non-Violent | 0.71 | 0.62 | 0.67 |
| | Violent | 0.87 | 0.91 | 0.89 |
| VGG Bi-LSTM | Non-Violent | 0.62 | 0.62 | 0.62 |
| | Violent | 0.86 | 0.86 | 0.86 |
| C3D | Non-Violent | 0.60 | 0.75 | 0.67 |
| | Violent | 0.90 | 0.82 | 0.86 |
| CNN-Transformer | Non-Violent | 0.83 | 0.67 | 0.74 |
| | Violent | 0.72 | 0.87 | 0.79 |

TABLE II
MODEL CLASSIFIER OUTPUT QUALITY SCORES

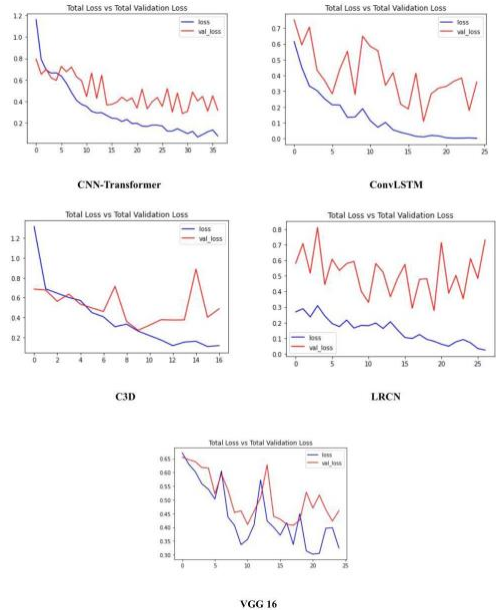

Fig. 3. Train Validation Loss History

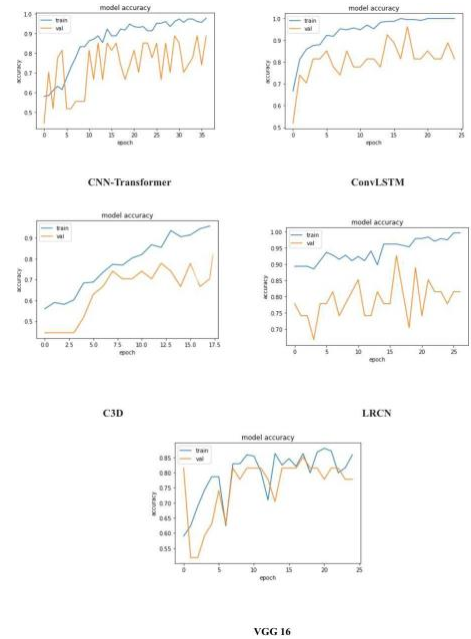

Fig. 4. Train Validation Accuracy History

*2) Model history:* The figure (3,4) represent the models train validation and test accuracy and loss history.

*3) CNN Filter Visualization:* The purpose of this study was to understand how the Convnets in the VGG-16 model used the filters on the 1st,3rd,8th 10th layers to learn the spatial information on a particular image.The convolution filters detect edges and lines and and forms a separated representation of the frames . The brightened up areas of the images are feature maps that are passed on to the top layers so the models can learn the useful information information

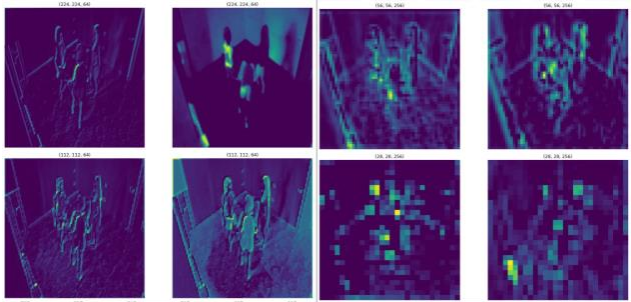

Fig. 5. Convolutional Filter on a Single Frame

*4) Confusion Matrices:* In figure (6), (0,0) represents true positives for non violent class ;(0,1) false positives for non-violent class; (1,0) false positives for violent class and (1,1) true positives for violent class. The confusion matrix for CNN-transformer shows that it can successfully identify 23 out of 30 cases, ConvLSTM identifies 24 cases successfully, LRCN and VGG-Bi-LSTM identifies 25 and 21 cases respectively and C3D identifies 23 cases in total. The models falsely identify the rest of the cases.

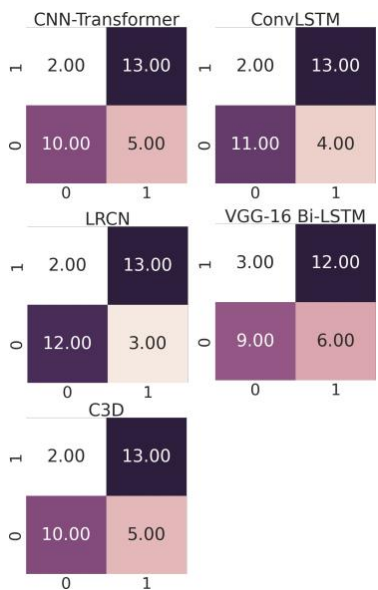

Fig. 6. Confusion matrices of the models

## VI. CONCLUSION AND FUTURE PLAN

The importance of surveillance systems has been reduced due to the inability to notify timely after detecting a crime. To eradicate this issue, we have proposed models that can help prevent crime by tracking and analyzing the footage from CCTV cameras in real time. Moreover, even if a security officer misses a notification at the moment, the investigation team can use the time at which the notification was sent afterward and find the recording of that exact time. This will make the job of the investigating team easier and help in prevailing justice for the victims. We aim to optimize our models by introducing more samples for the nonviolent class. After optimization, our task is to deploy our deep-learning models in a Jetson nano compared to a Raspberry Pi because it has a higher ram capacity, better CPU and 128 Cuda cores which will be very useful for running the models.